\pdfoutput=1

\documentclass[11pt]{article}

\usepackage{ACL2023}

\usepackage{times}
\usepackage{latexsym}
\usepackage{booktabs}
\usepackage{graphicx}
\usepackage{multirow}
\usepackage{multicol}
\usepackage{tikz}
\usetikzlibrary{shapes}
\usetikzlibrary{arrows.meta}
\usepackage{subfigure}
\usepackage{adjustbox}
\usepackage{amsmath}
\usepackage{listings}

\usepackage[T1]{fontenc}

\usepackage[utf8]{inputenc}

\usepackage{microtype}

\usepackage{inconsolata}

\lstnewenvironment{amrcode}[1][]
{
    \lstset{
        language=,
        basicstyle=\small\ttfamily,
        numbers=left,
        numberstyle=\tiny,
        stepnumber=1,
        numbersep=5pt,
        tabsize=2,
        extendedchars=true,
        breaklines=true,
        keywordstyle=\color{blue},
        showspaces=false,
        showtabs=false,
        showstringspaces=false,
        captionpos=b,
        frame=lines,
        framerule=0.5pt,
        rulecolor=\color{black},
        #1
    }
}
{}

\title{AMRs Assemble! 

Learning to Ensemble with Autoregressive Models for AMR Parsing}

\author{Abelardo Carlos Mart\'inez Lorenzo$^{1,2*}$ \qquad Pere-Llu\'is Huguet Cabot$^{1,2}$\thanks{$^*$ Equal contributions.} \\  \bf{Roberto Navigli$^2$}\\
         $^1$ Babelscape, Italy \\
         $^2$ Sapienza NLP Group, Sapienza University of Rome \\
         \texttt{\{martinez,huguetcabot\}@babelscape.com} \\
         \texttt{navigli@diag.uniroma1.it}}

\begin{document}
\maketitle
\begin{abstract}

In this paper, we examine the current state-of-the-art in AMR parsing, which relies on ensemble strategies by merging multiple graph predictions. Our analysis reveals that the present models often violate AMR structural constraints. To address this issue, we develop a validation method, and show how ensemble models can exploit SMATCH metric weaknesses to obtain higher scores, but sometimes result in corrupted graphs. Additionally, we highlight the demanding need to compute the SMATCH score among all possible predictions. To overcome these challenges, we propose two novel ensemble strategies based on Transformer models, improving robustness to structural constraints, while also reducing the computational time. Our methods provide new insights for enhancing AMR parsers and metrics.  Our code is available at   \href{https://www.github.com/babelscape/AMRs-Assemble}{github.com/babelscape/AMRs-Assemble}.

\end{abstract}

\section{Introduction}

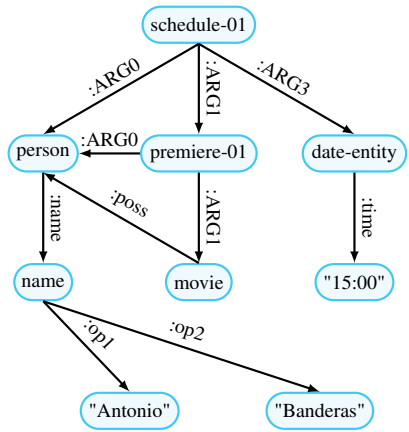
\begin{figure}[!t]
\centering
\resizebox{0.7\columnwidth}{!}{
\begin{tikzpicture}[blue/.style={rounded rectangle, draw=cyan!60, fill=cyan!5, very thick, minimum size=7mm, font=\fontsize{20}{20}},]
	\node[blue](z0) at (6.0,10.0) {schedule-01};
	\node[blue](z1) at (3.0,7.5) {person};
	\node[blue](z3) at (6.0,7.5) {premiere-01};
	\node[blue](z5) at (9.0,7.5) {date-entity};
	\node[blue](x4) at (9.0,5.0) {"15:00"};
	\node[blue](z2) at (3.0,5.0) {name};
	\node[blue](z4) at (6.0,5.0) {movie};
	\node[blue](x2) at (4.666666666666667,2.5) {"Antonio"};
	\node[blue](x3) at (8.333333333333334,2.5) {"Banderas"};
	\draw[->, very thick, -latex, scale=2] (z2.south) -- (x2.north) node[midway, above, sloped] {:op1};
	\draw[->, very thick, -latex, scale=2] (z2.south) -- (x3.north) node[midway, above, sloped] {:op2};
	\draw[->, very thick, -latex, scale=2] (z5.south) -- (x4.north) node[midway, above, sloped] {:time};
	\draw[->, very thick, -latex, scale=2] (z0.south) -- (z1.north) node[midway, above, sloped] {:ARG0};
	\draw[->, very thick, -latex, scale=2] (z1.south) -- (z2.north) node[midway, above, sloped] {:name};
	\draw[->, very thick, -latex, scale=2] (z0.south) -- (z3.north) node[midway, above, sloped] {:ARG1};
	\draw[->, very thick, -latex, scale=2] (z3.west) -- (z1.east) node[midway, above, sloped] {:ARG0};
	\draw[->, very thick, -latex, scale=2] (z3.south) -- (z4.north) node[midway, above, sloped] {:ARG1};
	\draw[->, very thick, -latex, scale=2] (z4.north) -- (z1.south) node[midway, above, sloped] {:poss};
	\draw[->, very thick, -latex, scale=2] (z0.south) -- (z5.north) node[midway, above, sloped] {:ARG3};
\end{tikzpicture}
}
 

\caption{AMR graph of the sentence: \textit{"Antonio Banderas scheduled the premiere of his movie at 3 pm"}.}
\label{fig:example}
\end{figure}

Semantic Parsing is the subfield of Natural Language Understanding \cite{navigli-2018} that aims to encode the meaning of a sentence in a machine-interpretable structure. One of the formalisms that has gained more attention is the Abstract Meaning Representation \cite[AMR]{banarescu-etal-2013-abstract}, which embeds the semantics of a sentence in a directed acyclic graph. In AMR, concepts are represented by nodes, and semantic relations between concepts by edges (see Figure \ref{fig:example}). AMR parsing has been applied to various areas of NLP, including Question Answering \citep{lim-etal-2020-know, bonial-etal-2020-infoforager, kapanipathi-etal-2021-leveraging}, Text Summarization \citep{hardy-vlachos-2018-guided,liao-etal-2018-abstract}, Information Extraction \citep{rao-etal-2017-biomedical}, and Machine Translation \citep{song-etal-2019-semantic}, and has been extended to non-English languages~\citep{anchieta-pardo-2020-semantically, blloshmi-etal-xlamr, oral-eryigit-2022-amr, bmr-etal-2022-bmr, martinez-lorenzo-etal-2022-fully}.

Current  AMR parsing approaches are based on Transformer sequence-to-sequence (seq2seq) models  \cite[SPRING]{bevilacqua-etal-2021-one}, which translate text into a linearized representation of the AMR graph. Recently, there have been some improvements through techniques such as pre-training on structural graph information \cite{bai-etal-2022-graph}, incorporating shallow semantic information \cite{chen-etal-2022-atp}, modifying ancestor information during decoding \cite{yu-gildea-2022-sequence}, and adding a structural graph prediction task during training \cite{cheng-etal-2022-bibl}. Nevertheless, in an attempt to push SMATCH~\cite{cai-knight-2013-SMATCH} performance, there has been a recent trend towards ensemble models, which merge AMR graph predictions from multiple systems. Some examples include Graphene~\cite{neurips-graphene-2021}, a graph mining algorithm that searches for the largest common structure among the graph predictions, or the Maximum Bayes SMATCH Ensemble~\cite{lee-etal-2022-maximum}, which introduces a Bayesian ensemble approach in order to create high-quality silver data. However, notwithstanding their higher performance, ensemble models are potentially  more vulnerable to producing corrupted AMR graphs. For instance, \citet{opitz-frank-2022-better} highlighted that better SMATCH scores do not always correlate with better parsing.


In this study, we conduct an investigation into the reasons why ensemble models improve their performance and in which cases they do so despite producing corrupted output. Our analysis reveals three significant drawbacks in these approaches: \textit{i)} ensemble systems do not consider structural constraints in AMR, treating AMR graphs as regular sets of triplets, \textit{ii)} they rely on SMATCH, which does not impose AMR constraints, exacerbating the problem of corrupted AMR graphs produced by ensemble methods that prioritize a higher score over adherence to structural constraints, as is the case with Graphene, and \textit{iii)} they are computationally expensive. Our findings highlight the need for more robust evaluation metrics that hold to the structural constraints of AMR.

\begin{table*}[!t]
    \centering
    \resizebox{\textwidth}{!}{

    \begin{tabular}{ccccc}
        \toprule
        \textbf{Phase} & \textbf{Task} & \textbf{Input}  & \textbf{Output} \\
        \midrule
        \midrule

        \multirow{5}{*}{\rotatebox[origin=c]{90}{Pre-training}} &  $\overline{t}\hat{g}2g$ & \texttt{<s>} $[mask]$  \texttt{<g>} $g_1, ... [mask]..., g_n$  \texttt{</s>} & \texttt{<s>} $g_1, g_2, ..., g_n$ \texttt{</s>} \\
        & $t\hat{g}2g$ & \texttt{<s>} $x_1, x_2, ..., x_m$  \texttt{<g>} $g_1, ... [mask]..., g_n$  \texttt{</s>} & \texttt{<s>} $g_1, g_2, ..., g_n$ \texttt{</s>}\\
        & $\hat{t}\hat{g}2g$ & \texttt{<s>} $x_1, ... [mask]..., x_m$  \texttt{<g>} $g_1, ... [mask]..., g_n$  \texttt{</s>}  & \texttt{<s>} $g_1, g_2, ..., g_n$ \texttt{</s>}\\
        & $\overline{t}\hat{g}_{1}...\hat{g}_{k}2g$ & \texttt{<s>} $[mask]$  \texttt{<g>} $g_{1}, ... [mask]_1..., g_{n}$ \texttt{<g>} ... \texttt{<g>} $g_{1}, ... [mask]_k..., g_{n}$   \texttt{</s>} & \texttt{<s>} $g_1, g_2, ..., g_n$ \texttt{</s>} \\
        & $\hat{t}\hat{g}_{1}...\hat{g}_{k}2g$ & \texttt{<s>} $x_1, ... [mask]..., x_m$   \texttt{<g>} $g_{1}, ... [mask]_1.., g_{n}$ \texttt{<g>} ... \texttt{<g>} $g_{1}, ... [mask]_k..., g_{n}$   \texttt{</s>} & \texttt{<s>} $g_1, g_2, ..., g_n$ \texttt{</s>} \\
        \midrule

        \multirow{3}{*}{\rotatebox[origin=c]{90}{Fine-tun.}} & $t\overline{p}_{1...k}2g$ & \texttt{<s>}  $x_1, x_2, ..., x_m$  \texttt{<g>} $[mask]$ \texttt{</s>} & \texttt{<s>} $g_1, g_2, ..., g_n$ \texttt{</s>} \\
        & $\overline{t}p_{1...k}2g$ & \texttt{<s>} $[mask]$  \texttt{<g>} $p^1_{1}, p^1_{2},..., p^1_{l_1}$ \texttt{<g>} ... \texttt{<g>} $p^k_{1}, p^k_{2},.., p^k_{l_k}$   \texttt{</s>} & \texttt{<s>} $g_1, g_2, ..., g_n$ \texttt{</s>} \\
        & $tp_{1...k}2g$ & \texttt{<s>}  $x_1, x_2, ..., x_m$  \texttt{<g>} $p^1_{1}, p^1_{2},..., p^1_{l_1}$ \texttt{<g>} ... \texttt{<g>} $p^k_{1}, p^k_{2},.., p^k_{l_k}$    \texttt{</s>} & \texttt{<s>} $g_1, g_2, ..., g_n$ \texttt{</s>} \\

        \bottomrule

    \end{tabular}
    }
    
    \caption{Pre-training and fine-tuning tasks. $t$ denotes text, $g$ denotes graph, $p$ denotes prediction.}
    \label{tab:tasks}
\end{table*}

In this paper, we present two novel ensemble strategies that address the above limitations of current approaches. In the first strategy, we follow previous \textit{merging} methods, showing how to train a seq2seq model to combine different predictions by taking into account both the original sentence and predictions from multiple models. In our second approach, we propose using \textit{selection} as the ensembling strategy, where we select the best graph instead of merging. Specifically, we base our method on the perplexity score of the model. Additionally, we propose a graph algorithm that checks the structural constraints in AMR graphs. Through these contributions, we aim to provide more robust and efficient solutions for ensembling AMRs.

\section{AMRs Assemble!} \label{model}

The task of AMR parsing can be framed as a seq2seq task, where the input $t = [x_1, x_2, ..., x_m]$ is a sequence of $m$ tokens and the output $g = [g_1, g_2, ..., g_n]$ is a linearized graph with $n$ tokens. To illustrate, the linearized representation of the AMR graph in Figure \ref{fig:example} is as follows:

\begin{lstlisting}[basicstyle=\scriptsize\ttfamily]
(z0 / schedule-01
    :ARG0 (z1 / person
        :name (z2 / name
            :op1 "Antonio"
            :op2 "Banderas"))
    :ARG1 (z3 / premiere-01
        :ARG0 z1
        :ARG1 (z4 / movie
            :poss z1)
    :time (z5 / date-entity
        :time "15:00")))
\end{lstlisting}

The goal of seq2seq AMR parsing task is to learn a function that models the conditional probability:
\begin{equation}
    p(g|x) = \prod_{t=1}^{n}p(e_t|e_{<t}, x),
\end{equation}
where $e_{<t}$ are the tokens of the linearized graph $g$ before step $t$. 

In this work, we use LongT5 \cite{guo-etal-2022-longt5} as the seq2seq model, which is specialized for long sequences, making it feasible to provide sentences and linearized graphs as input. 

\subsection{Pre-training} \label{pre-training}

To enhance the structure awareness of the language model in relation to AMR graphs and ensembling techniques, we extend the graph self-supervised pre-training method proposed by~\citet[AMRBart]{bai-etal-2022-graph}. Formally, we denote a sentence as $t = [x_1, x_2, ..., x_m]$, a graph as $g = [g_1, g_2, ..., g_n]$, and a prediction by system $s$ as $p_s = [p^s_{1}, p^s_{2}, ..., p^s_{l_s}]$. We follow AMRBart noise function with a dynamic masking rate and denote the noisy text and graph as $\hat{t}$ and $\hat{g}$, respectively. Moreover, let $\overline{t}$, $\overline{g}$, and $\overline{p}$ be the empty text, graph and prediction, respectively. As shown in Table \ref{tab:tasks}, our pre-training procedure includes tasks presented by AMRBart, such as: $i)$  empty text graph denoising ($\overline{t}\hat{g}2g$), $ii)$ text augmented graph denoising ($t\hat{g}2g$), and $iii)$ noisy text augmented graph denoising  ($\hat{t}\hat{g}2g$). Additionally, we introduce: $iv)$ empty text multiple graph denoising $\overline{t}\hat{g}_{1}...\hat{g}_{k}2g$, where the target graph is generated using different graphs' masked versions, and $v)$ noisy text augmented multiple graph denoising $\hat{t}\hat{g}_{1}...\hat{g}_{k}2g$, where we also include the masked sentence.

\subsection{Fine-tuning} \label{fine-tuning}

\paragraph{Prediction Corpus} To fine-tune ensemble systems we create a corpus of multiple predictions, starting from AMR 3.0 (LDC2020T02), which consists of 59,255 human-annotated sentence-graph pairs. We create five distinct train-test splits of this dataset in such a way that each test set is one fifth of the data and mutually exclusive. We train five separate models, based on \citet{blloshmi-etal-2021-spring}, on the corresponding training sets and use each model to generate predictions for its respective test set. By combining all of the predicted test sets, we obtain a corpus of AMR predictions. However, to train an ensemble model, it is necessary to merge predictions from multiple models. Therefore, we generate five distinct prediction corpora by repeating this process four additional times with different train-test split sets.

\paragraph{Strategy} 

Having obtained a corpus comprising multiple AMR predictions, we design a set of tasks that fine-tune the model for ensembling. The first task is AMR parsing ($t\overline{p}_{1...k}2g$), i.e., an AMR graph $g$ is generated by using only a sentence $t$ as input. In the second task, ensemble AMR predictions ($\overline{t}p_{1...k}2g$), the model is provided with a random set of AMR predictions $p$ without the corresponding sentence, so it is forced to use just graph information to ensemble. In the last task, ensemble AMR predictions using the sentence ($tp_{1...k}2g$), the model is provided with both a random set of AMR predictions $p$ and the original sentence $t$. To ensure that the model is able to learn to merge a variety of predictions, we randomly modify the samples by changing the order and number of predictions in each epoch. As a result of this process, we obtain a model that is able to effectively integrate information from multiple sources to generate high-quality AMR graphs, without relying on the expensive SMATCH metric as has been the case for previous ensemblers.

\subsection{Assemble! zero \& avg} \label{selector}

Nevertheless, using large autoregressive models to generate AMR graphs can be computationally expensive. Therefore, we propose an alternative approach that is more effective than previous merging strategies. Our method selects the best graph from a set of predictions. To achieve this, we introduce two novel scoring functions, in which we provide each predicted graph to the decoder of a model and extract their perplexity score, which can be done with a single forward pass. In the first method (Assemble!$_{zero}$), we leverage our trained ensemble model by providing the sentence and all the predictions in order to extract their perplexities and select the smallest one, i.e., we select prediction $p_{s'}$, where:
$$s' = \operatorname*{argmin}_{s\in \{ 1, ..., l \}}  perplexity(tp_{1...l}2p_s).$$ 
In the second method (Assemble!$_{avg}$), instead of using our ensembler, we use each model that generated the predictions to extract the perplexity for all the candidates. The final output is the graph $p_{s'}$ with the lowest average perplexity score, where:
$$s' = \operatorname*{argmin}_{s\in \{ 1, ..., l \}}  \frac{1}{l}\sum_{j\in \{ 1, ..., l \}} perplexity_j(t2p_s).$$

\begin{table*}[hbt!]
\resizebox{\textwidth}{!}{
\begin{tabular}{lccc|c|cc|cccccccc}
\toprule
& \textbf{Model} & \multicolumn{1}{l}{\textbf{Time (s)}} & \textbf{Corrupt.} & \textbf{SMATCH} &  \textbf{S2MATCH} &  \textbf{WWLK} & \textbf{Unlab.} & \textbf{NoWSD} & \textbf{Conc.}  & \textbf{NER}  & \textbf{Neg.}  & \textbf{Wiki} & \textbf{Reent.} & \textbf{SRL}  \\ \midrule \midrule
\multicolumn{1}{c}{\multirow{5}{*}{\rotatebox[origin=c]{90}{Predictions}}} & SPRING$_1$         & --- & ~~54 & 83.1   & 84.3 & 84.9         & 86.2            & 83.6           & 89.3           & 87.7          & 70.9         & 81.5           & 72.9          & 81.8          \\
\multicolumn{1}{c}{}                         & SPRING$_2$            &  ---     &  ~~52                                    & 82.7     &  83.9 & 82.1    & 85.9           & 83.2           & 89.0           & 87.5         & 72.6        & 80.2           & 73.0          & 81.4          \\
\multicolumn{1}{c}{}                         & SPRING$_3$        & ---   & ~~73                                    & 83.0      & 84.3 & 85.1     & 86.3            & 83.5           & 89.3           & 87.6          & 72.6         & 81.8           & 73.1 & 81.7     \\
\multicolumn{1}{c}{}                         & SPRING$_4$    & ---    & ~~33                                    & 82.8  & 84.0 & 84.1   & 86.0            &  83.3           & 88.9           & 87.3          & 71.7         & 81.5           & 72.8 & 81.4      \\
\multicolumn{1}{c}{}                         & SPRING$_5$          & ---    & ~104                                    & 82.6  & 83.9 & 84,5  &  85.8      &  83.2     & 89.2     & 87.3           & 73.0          & 81.6         & 73.0          & 81.4     \\ \midrule

 & Best$_{graph}$         & ---       & ~~51    & 86.5      & 87.5 & 88.0    & 89.0           & 86.9           & 91.7           & 89.9          & 76.5         & 83.8          & 77.7 & 85.4       \\ \midrule \midrule

\multirow{3}{*}{\rotatebox[origin=c]{90}{Mergers}}      
    & Graphene$_{base}$          & ~~~~~810        & 374                               & 83.6   & 84.8 & 84.9       & 86.6            & 84.1          & 89.8           & 88.0         & 73.5         & 81.2           & 72.3 & 82.4          \\
     & Graphene$_{SMATCH}$        & 11,884          & 260                            & \textbf{83.8}   & \textbf{85.0} & 85.0   & 86.9            & \textbf{84.4}           & \textbf{89.9}       & 88.1         & \textbf{73.8}         & 81.3           & 73.7 & \textbf{82.6}        \\
    & \textbf{Assemble!}          &  ~~~~~\textbf{431}         & ~~~\textbf{6}        & \textbf{83.8}  & \textbf{85.0} & \textbf{85.2}   &  \textbf{87.0}            &   84.3           & 89.7           & \textbf{88.3}         & 72.9         & \textbf{81.7}          & \textbf{74.2} & 82.3   \\ \midrule \midrule

\multirow{3}{*}{\rotatebox[origin=c]{90}{Selectors}}                   & SMATCH$_{avg}$         & ~~~~~493    & ~~51                                 & 83.7  & 85.0 & 85.3    & 86.8            & 84.2           & 89.7           & 88.1          & 73.3         & 82.0          & 73.9 & 82.4      \\
     & \textbf{Assemble!}$_{zero}$        & ~~~~~\textbf{256}          & ~~\textbf{13}                            & 83.9  & 85.1 & \textbf{85.4}     & 87.1            & 84.4           & \textbf{89.9}       & \textbf{88.3}         & \textbf{74.0}         & \textbf{82.2}           & 74.3 & 82.5        \\ 
    & \textbf{Assemble!}$_{avg}$            & ~~~~~635        & ~~22                               & \textbf{84.1}   & \textbf{85.3} & 84.4       & \textbf{87.2}           & \textbf{84.6}          & \textbf{89.9}          & \textbf{88.3}         & 73.3         & \textbf{82.2}           & \textbf{74.6} & \textbf{82.8}          \\ \bottomrule
\end{tabular}}
\caption{Results in AMR 3.0 test set. Bold indicates best. Columns: Model, computational time, corrupted graphs, SMATCH, S$^{2}$MATCH, WWLK and SMATCH breakdown. Row Blocks: Predictions, Best predicted and models.}
\label{table:overall-test-amr}
\end{table*}

\section{Experiments}

\subsection{Setup}

\textbf{~Dataset} \quad We evaluate our model using \text{AMR 3.0}. For pre-training, we use the same $200k$ silver data parsed by \citet[SPRING]{bevilacqua-etal-2021-one} from the Gigaword \textit{(LDC2011T07)} corpus. For fine-tuning, we use the corpus described in Section \ref{fine-tuning}.   

\textbf{Metric} \quad To evaluate our results, we employ the SMATCH metric, which quantifies the similarity between graphs by measuring the degree of overlap between their triplets, and SMATCH's breakdown metrics (see Appendix \ref{appendix:metric}). In addition, we validate our results using two novel AMR metrics: S$^{2}$MATCH \cite{opitz-etal-2020-amr} and WWLK \cite{opitz-etal-2021-weisfeiler}, in its WWLK-k3e2n version introduced in \citet{opitz-etal-2021-weisfeiler}.
 
\textbf{Ensemble Baselines} \quad  For our selection strategy, we use the system of \citet{barzdins-gosko-2016-riga} as a baseline, which calculates the average SMATCH score for a given graph in comparison to all the other candidates and selects the one with the highest score. 

Our baseline for merging is Graphene~\cite{neurips-graphene-2021}, an algorithm that identifies the graph with the most nodes and edges in common among different graphs. Specifically, given a pivot graph $g_{i}$ (where $i = 1, 2, ... , k$), Graphene collects votes from the other graphs for every existing vertex and existing/non-existing edges to correct $g_{i}$. We use two variants of Graphene, i) Graphene$_{base}$, where every input graph is chosen as a pivot graph once, and the best among the modified pivot graphs is chosen as the final prediction based on average support; and ii) Graphene$_{smatch}$, which is similar to Graphene$_{base}$ but chooses the best modified pivot graph based on average SMATCH score, similar to \citet{barzdins-gosko-2016-riga}. 

We do not compare our approach using Maximum Bayes SMATCH Ensemble~\cite{lee-etal-2022-maximum}, as it is a technique for producing high-quality silver data by combining SMATCH-based ensembling techniques with ensemble distillation, and its code and data are not publicly available.

\textbf{Our Models} \quad We simulate an ensemble of five models obtained by training SPRING on five different seeds, and apply these models to the test split of AMR 3.0 using each of them. Assemble! and Assemble!$_{zero}$ rely on LongT5 \cite{guo-etal-2022-longt5} and are trained as explained in Section \ref{model}.
\subsection{Results}

We present our results in Table \ref{table:overall-test-amr}. The \textit{Predictions} block shows the performance of each individual system used for ensembling, which have an average SMATCH score of 82.8. The \textit{Best$_{graphs}$} row portrays the upper bound of the selection strategy, where the SMATCH score is calculated with an oracle that selects the graph with the highest SMATCH. This score is 3.4 points higher than the best predictions. The \textit{Mergers} block presents the results of the ensembling strategies that combine predictions, where we observe that our model performs comparably to Graphene$_{smatch}$ but is 10 times faster. Furthermore, the \textit{Selector} block presents the results of the three different selection strategies, where the best graph is chosen out of a set of predictions. Our strategy outperforms SMATCH$_{avg}$ by 0.4 points while having a similar computation time. These results demonstrate the effectiveness of our proposed ensembling approaches and suggest that they may be an alternative to traditional merging methods.

\begin{figure*}[!t]
\centering
\subfigure[Gold AMR]{
\resizebox{0.49\columnwidth}{!}{
\begin{tikzpicture}[blue/.style={rounded rectangle, draw=cyan!60, fill=cyan!5, very thick, minimum size=7mm, font=\fontsize{20}{20}},]
	\node[blue](z0) at (6.0,10.0) {schedule-01};
	\node[blue](z1) at (3.0,7.5) {person};
	\node[blue](z3) at (6.0,7.5) {premiere-01};
	\node[blue](z5) at (9.0,7.5) {date-entity};
	\node[blue](x4) at (9.0,5.0) {"15:00"};
	\node[blue](z2) at (3.0,5.0) {name};
	\node[blue](z4) at (6.0,5.0) {movie};
	\node[blue](x2) at (4.666666666666667,2.5) {"Antonio"};
	\node[blue](x3) at (8.333333333333334,2.5) {"Banderas"};
	\draw[->, very thick, -latex, scale=2] (z2.south) -- (x2.north) node[midway, above, sloped] {:op1};
	\draw[->, very thick, -latex, scale=2] (z2.south) -- (x3.north) node[midway, above, sloped] {:op2};
	\draw[->, very thick, -latex, scale=2] (z5.south) -- (x4.north) node[midway, above, sloped] {:time};
	\draw[->, very thick, -latex, scale=2] (z0.south) -- (z1.north) node[midway, above, sloped] {:ARG0};
	\draw[->, very thick, -latex, scale=2] (z1.south) -- (z2.north) node[midway, above, sloped] {:name};
	\draw[->, very thick, -latex, scale=2] (z0.south) -- (z3.north) node[midway, above, sloped] {:ARG1};
	\draw[->, very thick, -latex, scale=2] (z3.west) -- (z1.east) node[midway, above, sloped] {:ARG0};
	\draw[->, very thick, -latex, scale=2] (z3.south) -- (z4.north) node[midway, above, sloped] {:ARG1};
	\draw[->, very thick, -latex, scale=2] (z4.north) -- (z1.south) node[midway, above, sloped] {:poss};
	\draw[->, very thick, -latex, scale=2] (z0.south) -- (z5.north) node[midway, above, sloped] {:ARG3};
	\label{fig:amr}
\end{tikzpicture}
}}
\subfigure[Pred 1. SMATCH 80,0.]{
\resizebox{0.49\columnwidth}{!}{
\begin{tikzpicture}[blue/.style={rounded rectangle, draw=cyan!60, fill=cyan!5, very thick, minimum size=7mm, font=\fontsize{20}{20}},]
	\node[blue](z0) at (6.0,10.0) {schedule-01};
	\node[blue](z1) at (3.0,7.5) {person};
	\node[blue](z3) at (6.0,7.5) {premiere};
	\node[blue](z5) at (9.0,7.5) {date-entity};
	\node[blue](x4) at (9.0,5.0) {"15:00"};
	\node[blue](z2) at (3.0,5.0) {name};
	\node[blue](z4) at (6.0,5.0) {movie};
	\node[blue](x2) at (4.666666666666667,2.5) {"Antonio"};
	\node[blue](x3) at (8.333333333333334,2.5) {"Banderas"};
	\draw[->, very thick, -latex, scale=2] (z2.south) -- (x2.north) node[midway, above, sloped] {:op1};
	\draw[->, very thick, -latex, scale=2] (z2.south) -- (x3.north) node[midway, above, sloped] {:op2};
	\draw[->, very thick, -latex, scale=2] (z5.south) -- (x4.north) node[midway, above, sloped] {:time};
	\draw[->, very thick, -latex, scale=2] (z0.south) -- (z1.north) node[midway, above, sloped] {:ARG0};
	\draw[->, very thick, -latex, scale=2] (z1.south) -- (z2.north) node[midway, above, sloped] {:name};
	\draw[->, very thick, -latex, scale=2] (z0.south) -- (z3.north) node[midway, above, sloped] {:ARG1};
	\draw[->, very thick, -latex, scale=2] (z3.south) -- (z4.north) node[midway, above, sloped] {:mod};
	\draw[->, very thick, -latex, scale=2] (z3.west) -- (z1.east) node[midway, above, sloped] {:poss};
	\draw[->, very thick, -latex, scale=2] (z0.south) -- (z5.north) node[midway, above, sloped] {:ARG3};
	\label{fig:pred1}

\end{tikzpicture}
}}
\subfigure[Pred 2. SMATCH 88,9.]{
\resizebox{0.49\columnwidth}{!}{
\begin{tikzpicture}[blue/.style={rounded rectangle, draw=cyan!60, fill=cyan!5, very thick, minimum size=7mm, font=\fontsize{20}{20}},]
	\node[blue](z0) at (6.0,10.0) {schedule-01};
	\node[blue](z1) at (3.0,7.5) {person};
	\node[blue](z3) at (6.0,7.5) {premiere-01};
	\node[blue](z5) at (9.0,7.5) {date-entity};
	\node[blue](x4) at (9.0,5.0) {"3:00"};
	\node[blue](z2) at (3.0,5.0) {name};
	\node[blue](z4) at (6.0,5.0) {movie};
	\node[blue](x2) at (4.666666666666667,2.5) {"Antonio"};
	\node[blue](x3) at (8.333333333333334,2.5) {"Banderas"};
	\draw[->, very thick, -latex, scale=2] (z2.south) -- (x2.north) node[midway, above, sloped] {:op1};
	\draw[->, very thick, -latex, scale=2] (z2.south) -- (x3.north) node[midway, above, sloped] {:op2};
	\draw[->, very thick, -latex, scale=2] (z5.south) -- (x4.north) node[midway, above, sloped] {:time};
	\draw[->, very thick, -latex, scale=2] (z0.south) -- (z1.north) node[midway, above, sloped] {:ARG0};
	\draw[->, very thick, -latex, scale=2] (z1.south) -- (z2.north) node[midway, above, sloped] {:name};
	\draw[->, very thick, -latex, scale=2] (z0.south) -- (z3.north) node[midway, above, sloped] {:ARG1};
	\draw[->, very thick, -latex, scale=2] (z3.west) -- (z1.east) node[midway, above, sloped] {:ARG0};
	\draw[->, very thick, -latex, scale=2] (z3.south) -- (z4.north) node[midway, above, sloped] {:ARG1};
	\draw[->, very thick, -latex, scale=2] (z4.north) -- (z1.south) node[midway, above, sloped] {:poss};
	\draw[->, very thick, -latex, scale=2] (z0.south) -- (z5.north) node[midway, above, sloped] {:ARG3};
	\label{fig:pred2}

\end{tikzpicture}
}}
\subfigure[Graphene. SMATCH 85,0.]{
\resizebox{0.49\columnwidth}{!}{
\begin{tikzpicture}[blue/.style={rounded rectangle, draw=cyan!60, fill=cyan!5, very thick, minimum size=7mm, font=\fontsize{20}{20}},]
	\node[blue](z0) at (6.0,10.0) {schedule-01};
	\node[blue](z1) at (3.0,7.5) {person};
	\node[blue](z3) at (6.0,7.5) {premiere};
	\node[blue](z5) at (9.0,7.5) {date-entity};
	\node[blue](x5) at (7.5,5.0) {"15:00"};
	\node[blue](x6) at (9.5,5.0) {"3:00"};
	\node[blue](z2) at (3.0,5.0) {name};
	\node[blue](z4) at (4.5,5.0) {movie};
	\node[blue](x2) at (4.666666666666667,2.5) {"Antonio"};
	\node[blue](x3) at (8.333333333333334,2.5) {"Banderas"};
	\draw[->, very thick, -latex, scale=2] (z2.south) -- (x2.north) node[midway, above, sloped] {:op1};
	\draw[->, very thick, -latex, scale=2] (z2.south) -- (x3.north) node[midway, above, sloped] {:op2};
	\draw[->, very thick, -latex, scale=2] (z5.south) -- (x5.north) node[midway, above, sloped] {:time};
	\draw[->, very thick, -latex, scale=2] (z5.south) -- (x6.north) node[midway, above, sloped] {:time};
	\draw[->, very thick, -latex, scale=2] (z0.south) -- (z1.north) node[midway, above, sloped] {:ARG0};
	\draw[->, very thick, -latex, scale=2] (z1.south) -- (z2.north) node[midway, above, sloped] {:name};
	\draw[->, very thick, -latex, scale=2] (z0.south) -- (z3.north) node[midway, above, sloped] {:ARG1};
	\draw[->, very thick, -latex, scale=2] (z3.south) to[bend right] node[midway, above, sloped] {:mod} (z4.north);
	\draw[->, very thick, -latex, scale=2] (z4.north) -- (z1.south) node[midway, above, sloped] {:poss};
	\draw[->, very thick, -latex, scale=2] (z3.west)  to[bend right] node[midway, above, sloped] {:poss} (z1.east);
	\draw[->, very thick, -latex, scale=2] (z3.west)  to[bend left] node[midway, above, sloped] {:ARG0} (z1.east);
	\draw[->, very thick, -latex, scale=2] (z3.south) to[bend left] node[midway, above, sloped] {:ARG1} (z4.north);
	\draw[->, very thick, -latex, scale=2] (z0.south) -- (z5.north) node[midway, above, sloped] {:ARG3};
	\label{fig:graphene}

\end{tikzpicture}
}}
\caption{AMR representations of the sentence: \textit{"Antonio Banderas scheduled the premiere of his movie at 3 pm"}.}
\label{fig:all}
\end{figure*}
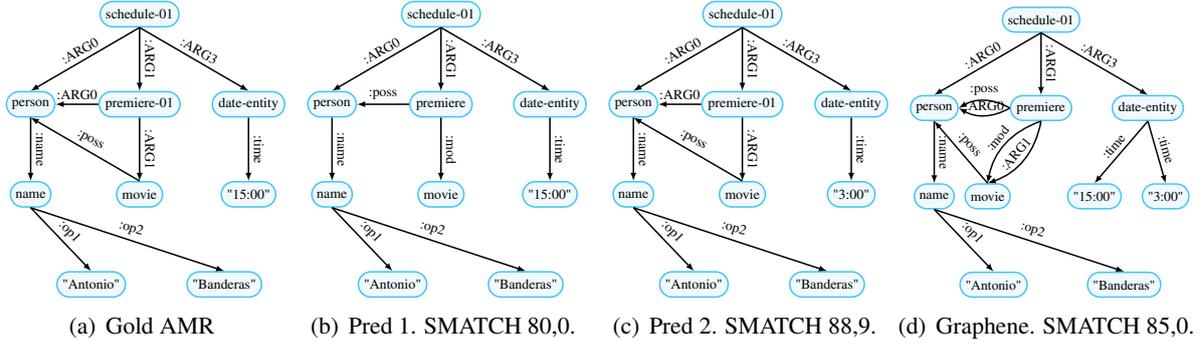

\subsection{Analysis}

While our model is able to effectively ensemble graphs or select the most accurate one from a set of predictions in an efficient and competitive manner, it is important to note that a higher SMATCH score does not always equate to the best graph if the graph has structural issues. This is because the SMATCH metric simply views the graph as a set of triplets. For example, the AMR graph illustrated in Figure \ref{fig:amr} is treated as the following triplets:

\begin{lstlisting}[basicstyle=\scriptsize\ttfamily]
(empty, :root, z0) ^ 
(z0, :instance, schedule-01) ^ 
(z0, :ARG0, z1) ^ 
(z1, :instance, person) ^ 
(z1, :name, z2) ^ 
(z2, :instance, name) ^ 
(z2, :op1, "Antonio") ^ 
(z2, :op2, "Banderas") ^ 
(z0, :ARG1, z3) ^ 
(z3, :instance, premiere-01) ^ 
(z3, :ARG0, z1) ^ 
(z3, :ARG1, z4) ^ 
(z4, :instance, movie) ^ 
(z4, :poss, z1) ^ 
(z0, :ARG3, z5) ^ 
(z5, :instance, date-entity) ^ 
(z5, :time, "15:00")
\end{lstlisting}
 
SMATCH calculates the degree of overlapping between two sets of triplets, but it does not consider the implicit AMR constraints. To address this problem, we develop an algorithm that checks some AMR violations in graphs: \textit{i)} non-predicate nodes with \texttt{:ARG} relations, \textit{ii)}  predicate nodes with \texttt{:op} or \texttt{:snt} relations, \textit{iii)}  compositional issues in entity structures, and \textit{iv)} compositional issues in connector structures. The \textit{Corrupt.} column in Table \ref{table:overall-test-amr} shows the number of graphs with structural problems out of $1898$ graphs. This highlights the limitation of previous ensemblers, such as Graphene, which do not consider these structural constraints.

\textbf{Ensembling} \quad As demonstrated in the \textit{Corrupt.} column of Table \ref{table:overall-test-amr}, our ensemble method  has a significantly lower number of graphs with structural issues  ($0.3$\%) as compared to Graphene$_{base}$ and Graphene$_{smatch}$ ($13.7$-$19.7$\%). This is because previous ensemble models 
are only focused on achieving a higher SMATCH metric, interpreting the graphs just as a set of triplets. 
This leads to ensembled graphs with violations of AMR guidelines and semantic inconsistencies. Figure \ref{fig:graphene} shows the Graphene$_{smatch}$ generated graph, and Figures \ref{fig:pred1} and \ref{fig:pred2} the two predictions used for ensembling. The Graphene$_{smatch}$ graph presents multiple AMR violations that are not in its predictions, e.g., the \textit{premiere} node is connected to the \textit{movie} node with two different relations because Graphene cannot decide which is the correct edge (both relations have the same probability), and one of the relations is an argument relation (i.e., \texttt{:ARG}), which cannot be used with non-predicate nodes since their meaning is encoded in PropBank frames. 


\textbf{SMATCH} \quad Graphene results in Table \ref{table:overall-test-amr}  are competitive despite having a higher percentage of structural issues in the ensembled graphs. This discrepancy can be attributed to the inherent properties of the SMATCH metric, which penalizes missing triplets more than wrong triplets. For example, the ensembled graph of Figure \ref{fig:graphene} obtains a higher SMATCH score than the prediction of Figure \ref{fig:pred1}, since, in case of doubt, selecting both triplets (relations  $ARG1$ and $mod$) from node \textit{premiere} to node \textit{movie} results in a higher score than selecting only the wrong triplet. This illustrates how current ensemble models exploit SMATCH weaknesses to attain higher scores. In contrast, our approaches provide competitive results while also being more robust to AMR constraints.

Furthermore, as highlighted in \citet{opitz-frank-2022-better}, the current scores of AMR parsers and ensemblers (around 0.83 and 0.84, respectively) are higher than the average annotator vs. consensus inter-annotator agreement reported in \citet{banarescu-etal-2013-abstract} (0.83 and 0.79 in newswire and web text, respectively). Additionally, WWLK results  in Table \ref{table:overall-test-amr} show how SPRING$_{3}$ predictions achieve comparable results to all ensemble models. Therefore, given the issues discussed above, the suitability of SMATCH for evaluating the model's performance beyond 0.83 has to be called into question.


\section{Conclusion}

In this paper, we leveraged self-supervised pre-training and a denoising autoencoder architecture to achieve strong results in merging AMR graph predictions. We also introduced two novel approaches for ensembling that select the best prediction from a set of candidates using simple and efficient perplexity score functions. These results suggest that the selection strategy is a promising alternative for ensembling, since it achieves competitive performance while being less expensive. 

Furthermore, we developed an algorithm that checks the structural AMR constraints in parsing outputs. This allowed us to perform an analysis that revealed how previous ensemble models produce higher score graphs but exploit SMATCH weaknesses that lead to increased structural issues. Overall, our findings highlight the need for more robust evaluation metrics and ensemble models that are designed to adhere to the structural constraints. 
\newpage
\section{Limitations}

Our proposed ensemble approach for training the Transformer architecture has demonstrated promising results for the task of AMR ensembling. However, there are limitations that warrant further investigation in future research.

Our first limitation is the lack of generalization, as the approach was only evaluated on AMR parsing. Therefore, the application of an autoregressive ensembling model has not yet been tested on other Natural Language Processing tasks. 

Moreover, in order to properly compare each ensemble system under the same conditions, we base all our experiments using the same underlying architecture, i.e. SPRING. There needs to be an exploration of these approaches using more recent, better performing parsers. However, this will require access to such systems.

Furthermore, the computational cost is also a limitation, as even though our proposed merger method, Assemble!, is more efficient than previous ensemblers, it is still computationally expensive, and particularly when we have to ensemble long graphs from multiple predictions. Moreover, as our Assemble! model is based on LongT5, it might be  challenged when working with large datasets or when running experiments on resource-constrained systems. Therefore, we encourage the use of ensembling strategies focused on selecting the best graphs instead of merging.

Lastly, as our ensemble approach is based on Transformer, results can be difficult to interpret, as it can be challenging to understand how the generated graph has been ensembled by different predictions, leading to a lack of interpretability.

In summary, the proposed ensemble approach for training the Transformer architecture has shown promising results for the task of AMR ensembling and has the potential to be applied to other tasks, however, further research is necessary to address its limitations and improve performance.

\section{Ethics Statement}

Regarding the ethical and social implications of our approach for AMR ensembling, we do not believe it could have a negative impact.  However, since ethical considerations are an important aspect of any research and development project, we will discuss a few ethical considerations here.

First, one potential concern is the use of Transformer-based models, which have been shown to perpetuate societal biases present in the data used for training. Our approach relies on the use of these models, and it is crucial to ensure that the data used for training is diverse and unbiased.

Second, it is important to consider the potential impact of the proposed ensemble strategies on marginalized communities. It is possible that these strategies may inadvertently perpetuate or amplify existing biases in the data used to train and test these systems. Therefore, it is important to ensure that the proposed ensemble strategies are tested on a diverse set of data and that any biases are identified and addressed.

In conclusion, the proposed ensemble strategies in this paper can potentially have positive impact on the field of AMR parsing, however, it is important to consider the ethical implications of this research and take steps to mitigate any potential negative consequences.

\section*{Acknowledgments}

\begin{center}
\noindent
\begin{minipage}{0.1\linewidth}
    \raisebox{-0.25\height}{\includegraphics[trim =0mm 5mm 5mm 5mm,clip,scale=0.045]{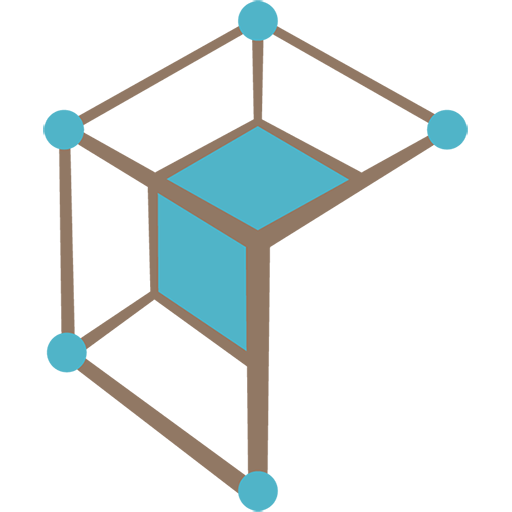}}

\end{minipage}
\hspace{0.005\linewidth}
\begin{minipage}{0.72\linewidth}
The authors gratefully acknowledge the support of the European Union’s Horizon 2020 research project \textit{Knowledge Graphs at Scale} (KnowGraphs) under the Marie  Marie Sk\l{}odowska-Curie grant agreement No \href{https://cordis.europa.eu/project/id/860801}{860801}.

  \vspace{1ex}
\end{minipage}
\hspace{0.005\linewidth}
\begin{minipage}{0.1\linewidth}
  \vspace{0.05cm}
\raisebox{-0.25\height}{\includegraphics[trim =0mm 5mm 5mm 5mm,clip,scale=0.060]{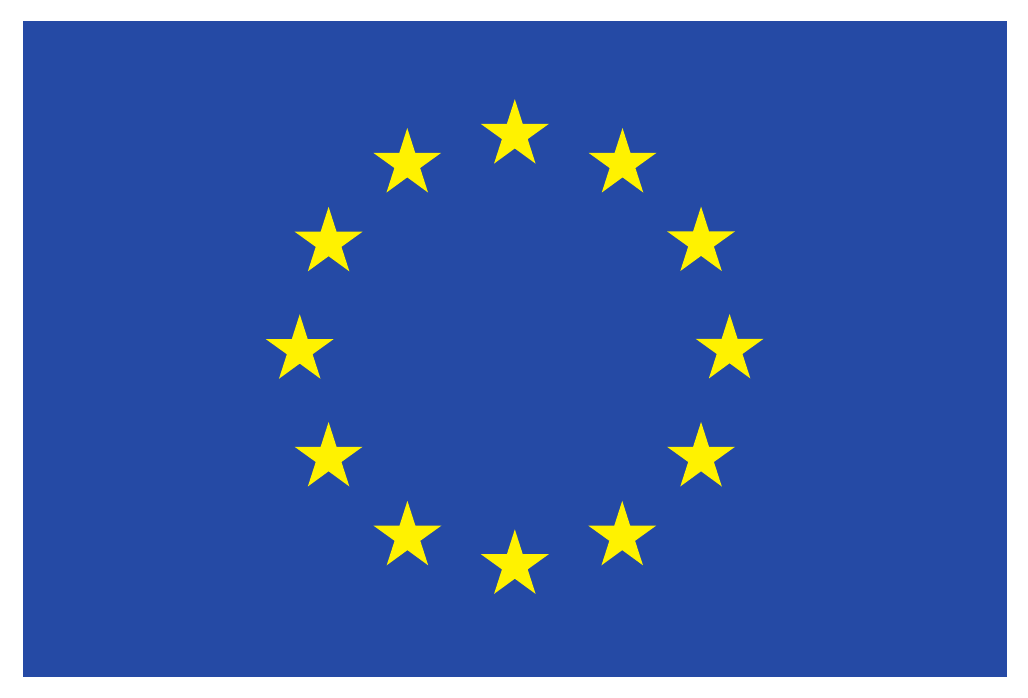}}
  \vspace{0.05cm}
\end{minipage}\\
\end{center}
The last author gratefully acknowledges the support of the PNRR MUR project PE0000013-FAIR. 
The authors sincerely thank Lorenzo Proietti and Stefano Perrella for their contribution to this project.

\bibliography{anthology,custom}
\bibliographystyle{acl_natbib}

\newpage
\appendix

\section{Model Hyper-Parameters}
Table \ref{table:hp-space} lists hyperparameters and search space for the experiments with SPRING models and our Assemble!. The masking probabilities of the pre-training task are: i) $\overline{t}\hat{g}2g$ -- 0.35\%, ii) $t\hat{g}2g$ -- 0.35\%, iii) $\hat{t}\hat{g}2g$ -- from 0.15\% to 0.85\% incrementing by epoch, iv) $\overline{t}\hat{g}_{1}...\hat{g}_{k}2g$ -- 0.55\%, and v) $\hat{t}\hat{g}_{1}...\hat{g}_{k}2g$ -- 0.55\%.

\label{sec:appendix}
\begin{table}[hbt!]
\centering
\resizebox{\columnwidth}{!}{
\begin{tabular}{lcc}
\toprule
\textbf{Group} & \textbf{Parameter} &  \textbf{Values} \\
\midrule
\multirow{7}{*}{\shortstack[l]{Pre-training}}
                       & Optimizer & Adafactor \\
                       & Batch size & 1 \\
                       & Dropout & 0.2 \\
                       & Attent. dropout & 0.0 \\
                       & Grad. accum. & 32 \\
                       & Weight decay & 0.01 \\    
                       & LR & 0.0001 \\
                       & LR sched. & Inverse sqrt \\
                       & Beamsize & 5 \\
\midrule
\multirow{7}{*}{\shortstack[l]{Fine-tuning}}
                       & Optimizer & Adafactor \\
                       & Batch size & 1.0 \\
                       & Dropout & 0.1 \\
                       & Attent. dropout & 0.0 \\
                       & Grad. accum. & 32.0 \\
                       & Weight decay & 0.01 \\    
                       & LR & 0.00001 \\
                       & LR & Constant \\
                       & Beamsize & 5 \\
\midrule
 \bottomrule
\end{tabular}}
\caption{Final hyperparameters and search space for the experiments.}
\label{table:hp-space}
\end{table}

\section{Hardware and size of the model}
We performed experiments on a single NVIDIA 3090 GPU with 64GB of RAM and Intel®
Core™ i9-10900KF CPU. The total number of trainable parameters of SKD is 434,883,596. The pre-training phase on the silver data requires 168 hours, whereas fine-tuning requires 216 hours.

\section{BLINK}
All systems from Table \ref{table:overall-test-amr} use BLINK~\cite{wu-etal-2020-scalable} for wikification. For this purpose, we used the $blinkify.py$ script from the SPRING repository.

\section{Metric}\label{appendix:metric}

To evaluate the predictions, we use the SMATCH metric and the extra scores of \citet{damonte-etal-2017-incremental}: \textit{i)} Unlabel, compute on the predicted graphs after removing all edge labels, \textit{ii)} No WSD, compute while ignoring Propbank senses (e.g., duck-01 vs duck-02), \textit{iii)} Wikification, F-score on the wikification (:wiki roles), \textit{iv)} NER, F-score on the named entity recognition (:name roles), \textit{v)} Negations, F-score on the negation detection (:polarity roles), \textit{vi)} Concepts, F-score on the concept identification task, \textit{vii)} Reentrancy, computed on reentrant edges only, \textit{viii)} Semantic Role Labeling (SRL), computed on :ARG-i roles only.

\section{Data}
The AMR 3.0 data used in this paper is licensed under the \textit{LDC User Agreement for Non-Members} for LDC subscribers, which can be found \href{https://catalog.ldc.upenn.edu/LDC2020T02}{here}.

\end{document}